\documentclass[letterpaper, 10 pt, conference]{ieeeconf}
\usepackage{subcaption}
\usepackage{graphicx}
\usepackage{multirow}
\usepackage{booktabs}
\usepackage{amsmath}
\usepackage{amsfonts}
\usepackage{caption}
\usepackage{subcaption}
\usepackage{hyperref}

\ifdefined\draft
    \usepackage{comment}
    \usepackage[textsize=footnotesize]{todonotes}
    \DeclareRobustCommand{\enrico}[1]
    {{\todo[color=green!40,inline]{EM: #1}}}

    \DeclareRobustCommand{\update}[1]
    {{\color{red} #1 \color{black}}} 
    
    \usepackage[nopar=false]{lipsum}
    
    \DeclareRobustCommand{\fake}[1]{\color{gray!40!}{\lipsum[][#1]}\color{black}}
    
\else 
\newcommand{\enrico}[1]{}
\newcommand{\nicola}[1]{}
\newcommand{\fake}[1]{}

\newcommand{\update}[1]{#1}
\usepackage{pbalance} 
\fi

\IEEEoverridecommandlockouts                    
\title{\LARGE \bf
A Robust Filter for Marker-less Multi-person Tracking in Human-Robot Interaction Scenarios

}

\author{Enrico Martini$^{1,2}$*, Harshil Parekh$^{2}$, Shaoting Peng$^{2}$, Nicola Bombieri$^{1}$, and Nadia Figueroa$^{2}$
\thanks{%
This work has been supported by the "PREPARE" project (n. F/310130/05/X56 - CUP: B39J23001730005) - D.M. MiSE 31/12/2021,
and the PNRR research activities of the consortium iNEST (Interconnected North-Est Innovation Ecosystem) funded by the European Union Next-GenerationEU (Piano Nazionale di Ripresa e Resilienza (PNRR) – Missione 4 Componente 2, Investimento 1.5 –D.D. 1058 23/06/2022, ECS 00000043). This manuscript reflects only the Authors’ views and opinions, neither the European Union nor the European Commission can be considered responsible for them.%
*Corresponding author: {\tt\small enrico.martini@univr.it}}
\thanks{$^{1}$Department of Engineering for Innovation Medicine, University of Verona, Verona, Italy.}%
\thanks{$^{2}$GRASP Lab, Department of Mechanical Engineering and Applied Mechanics, University of Pennsylvania, Philadelphia, USA.}%
}

\begin{document}

\maketitle
\thispagestyle{empty}
\pagestyle{empty}

\begin{abstract}
Pursuing natural and marker-less human-robot interaction (HRI) has been a long-standing robotics research focus, driven by the vision of seamless collaboration without physical markers. Marker-less approaches promise an improved user experience, but state-of-the-art struggles with the challenges posed by intrinsic errors in human pose estimation (HPE) and depth cameras. These errors can lead to issues such as robot jittering, which can significantly impact the trust users have in collaborative systems. We propose a filtering pipeline that refines incomplete 3D human poses from an HPE backbone and a single RGB-D camera to address these challenges, solving for occlusions that can degrade the interaction. Experimental results show that using the proposed filter leads to more consistent and noise-free motion representation, reducing unexpected robot movements and enabling smoother interaction.
\end{abstract}

\section{Introduction}
In recent years, integrating robots into homes, offices, and public spaces has become increasingly common, changing the dynamics of Human-Robot Interaction (HRI) and making it more important to ensure that these interactions remain safe and robust. The robot's knowledge of its environment is an essential component of this coexistence.

Traditional HRI tasks (e.g., collision avoidance~\cite{8932105} and teleoperations~\cite{CERULO201775}) have heavily relied on marker-based motion capture systems to detect human poses for their high precision and low latency. Such systems use reflective markers to accurately track human movements~\cite{5410185,lockwood2023gaussian}, detected by multiple fixed high-speed cameras, achieving speeds up to 250Hz with sub-millimeter errors~\cite{Kiss_2018}. 
However, the intrusiveness, limited mobility, and cost implications associated with marker-based motion capture systems preclude their integration into everyday human-robot interaction tasks at the consumer level. 

In contrast, marker-less motion tracking represents a leap towards more natural and unobtrusive human-robot interactions~\cite{Colyer2018}, providing a cheap and easy solution. 
Human pose estimation (HPE) is a computer vision task that involves determining the position of a person's joints from images or videos. Recent advances in deep learning algorithms have greatly improved performance and robustness, making it a rapidly evolving area of research~\cite{zheng2023deep}. Previous works~\cite{Du2014, LIU2018355} combined RGB-D cameras and HPE to interpret human motions in 3D. Still, performance falls short as errors increase along with the distance from the camera~\cite{s120708640}. In dynamic environments, where humans and robots coexist, there are even more limitations regarding accuracy, real-time processing, and handling occlusions or complex human poses~\cite{Colyer2018}. 
While some research~\cite{wei2021vision, Yasar2024,kothari2023enhanced} attempt to use marker-less human-pose detections, they employ multiple sensors~\cite{wang2023multimodal}. 
In the context of HRI, most approaches prefer marker-based solutions due to the quality of measurement~\cite{Lorenzini2023} and frequency of data which makes reactive motion policies possible~\cite{DBLP:journals/corr/abs-2201-10392}.
In~\cite{dagioglou2021smoothing, takamido2023learning}, authors used a single camera and an HPE to demonstrate the desired end-effector movement of a robotic arm.
Both approaches are designed for single-person and avoid simultaneous robot and human movement.
All these approaches do not consider missing data information and large errors, which often occur in real-world scenarios and can degrade the estimation results. 
A multi-person collaborative scenario is even more prone to error because the HPE may not detect body parts occluded by the robot, objects, and other people; interactions between people, overlapping body parts, and self-occlusions further degrade poses~\cite{martini2023denoising}.

\begin{figure*}[t!]
    \centering
\includegraphics[width=0.987\linewidth]{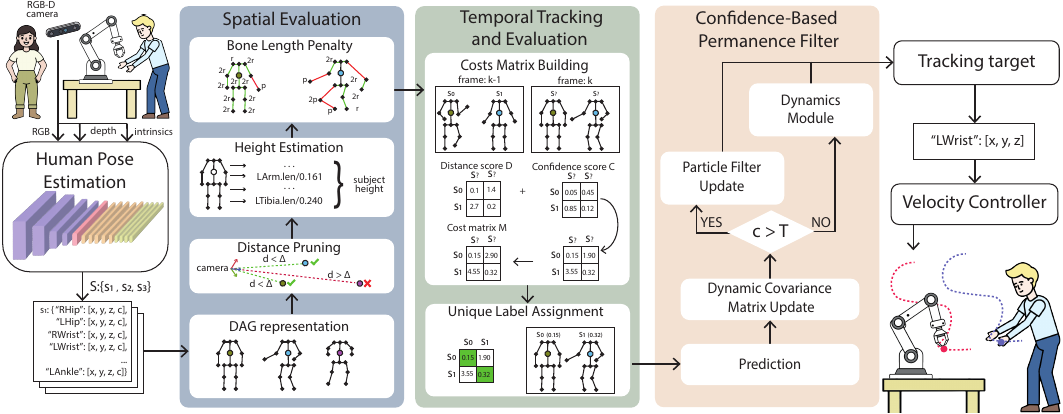}
    \caption{The proposed filter overview: a single camera sends the stream of a multi-person HRI scenario to a generic HPE backbone that outputs a set of 3D poses. The spatial evaluation node adjusts the initial prediction confidence based on the spatial relationship between keypoints. The temporal tracking and evaluation node assigns a unique label to each person in the scene, consistent across frames. The permanence filter uses previous trajectories to detect occlusions in the scene. This results in a refined set of skeletons fed into a tracking target node that provides the 3D position goal for the velocity controller.
    }
    \label{fig:pipeline}
\end{figure*}

To overcome these limitations, we propose a novel filtering pipeline that can provide a reliable target for the robot to handle multiple people in a close environment from a single camera. Since it only takes a set of 3D skeletons and the prediction confidence as input, it can be combined with any existing HPE backbone.
Figure~\ref{fig:pipeline} shows our filter, consisting of three steps: a spatial node, a temporal node, and a permanence filter node. The spatial node determines whether each prediction is reliable based on the spatial relationship between detected human joints. The temporal node tracks the people in the scene and assigns a unique label to each prediction based on the position of each body joint and the reliability of the prediction.
The filter node leverages the information from the previous nodes to detect occlusions and make smooth predictions even in prolonged and complete occlusions.

We tested the proposed pipeline in a multi-person scenario
on four tasks of increasing complexity with seven different operators. Each task, described in Figure~\ref{fig:tasks}, emulates real-world actions, such as inspections and handovers of light and heavy objects. 
The robot closely follows the operator's left wrist while another human collaborates with the operator, simulating the learning task from human-human interactions. 
We compared our filter with two Kalman filter variants that work in real-time and OpenPose~\cite{cao2017realtime}, assumed as a baseline. Comparing the results of these methods with a marker-based motion capture system used as ground truth shows that the proposed filter outperforms the other methods in all tasks. Furthermore, we monitored the safety distance between the robot and the operator's hand. We asked all study participants to rate on a Likert scale how safe and comfortable the interaction felt. Our method provided the safest and smoothest interaction with the least variability in the safety distance, surpassing traditional tracking algorithms. 
\update{
The code is publicly available at \href{https://github.com/penn-figueroa-lab/markerless-human-perception}{https://github.com/penn-figueroa-lab/markerless-human-perception}.
}


\section{Filtering pipeline}
We propose a plug-and-play filter that smoothes and tracks human poses estimated by any Multi-Person HPE to enable easier cooperation between humans and robots. Figure~\ref{fig:pipeline} shows an overview of the pipeline. A calibrated camera provides the RGB image, depth matrix, and intrinsics (i.e., focal length and principal points) to the HPE module. All HPE platforms rely on neural networks, but some infer the keypoints directly in 3D space~\cite{zhang2022voxeltrack}, while others provide 2D keypoints that are projected in 3D using the depth matrix~\cite{cao2017realtime,chen20232d}. Either way, the result is a set of skeletons $S_k$, where each skeleton $s_k \in S_k$ contains the 3D keypoints (i.e., coordinates of the human joints) extrapolated from an image, along with the confidence of the prediction. All those neural networks are trained on large generic datasets containing annotated images or videos of multiple people in various poses. When applied to an environment not present in the training dataset, such as an industrial plant, raw predictions are often inaccurate, with missing or misplaced keypoints, affecting HRI feasibility. Our solution to improve the outcome of generic backbones consists of a plug-and-play filter composed of a spatial evaluation module, a temporal evaluation and tracking module, and a permanence filter. The following sections describe each block in more detail.


\subsection{Spatial Evaluation}

This module determines whether to increase or decrease the confidence of each prediction based on the spatial relationship between keypoints, and the relationship  between the skeleton and the camera.
Each skeleton that enters this module is represented as a set of labeled 3D keypoints, each one defined as:
\begin{equation}
\text{kp}_{i,k} = [x,y,z,c]    
\end{equation}
where $x,y,z$ is the 3D position in the camera frame of keypoint $i$ at time $k$, and $c$ is the confidence of the prediction from the HPE.
To fully evaluate spatial properties, we build a direct acyclic graph (DAG), where each keypoint is a node, and each link between two adjacent keypoints is represented as an edge. Various HPE backbones produce different keypoints, so to make our method effective, we add keypoints at strategic locations to construct the DAG, if not present. For instance, consider a simple HPE that outputs only the 12 basic keypoints (shoulders, hips, elbows, wrists, knees, and ankles). We construct the central keypoint (i.e., the root) by averaging the position of the shoulders and hips. 

We then consider the Euclidean distance between the root of each skeleton and the camera position. The skeleton is discarded if the distance exceeds a certain threshold (i.e., $\varDelta$). This is because even the newest depth sensors are subject to error, and the signal-to-noise ratio decreases significantly as the distance from the camera increases. In addition, each neural network must resize the input images to make them the same size as those used during training. This often leads to hallucinations and general inaccuracy in detecting people at long distances.

The height of the corresponding human is then estimated for each skeleton. Biomechanics and anthropometry studies have shown a well-defined relationship between segment length and human height \cite{drillis1964body}. These segment proportions are a good approximation \cite{winter2009biomechanics}, so we use them to estimate the subject's height based on the distance between joints. For example, knowing that the forearm is $0.146$ times the height, if the measured distance between the elbow and wrist is $0.25$ m, the height is approximately $1.71$ m. Each arc of the DAG is used to derive a height hypothesis in the height estimation process. All possible heights are filtered and averaged to estimate a plausible height. 

Then, for each arc of the DAG, if its length is compatible with the estimated height, the confidence of its two keypoints is increased by a reward factor $r$. Conversely, if the arc's length is incorrect, the keypoint's confidence is decreased by a penalty factor $p$. If it is connected to two invalid bones, it is penalized by $2p$ because it's probably a false detection. On the other hand, if all bones connected to a keypoint are valid, the keypoint is likely to be correctly detected by the HPE, so its confidence increases by a $2r$ factor.

\subsection{Temporal Tracking and Evaluation}
This module assigns a unique label to each skeleton to keep track of the people in the scene. We solve the problem using the \textit{Hungarian method}~\cite{kuhn1955hungarian}, an optimization algorithm that is specifically designed for the assignment problem. Given a $n\times m$ cost matrix $M\in N^+$, where $n$ is the number of skeletons detected at time $k-1$ and $m$ is the number of skeletons detected at time $k$, the Hungarian method finds an assignment between the $n$ old skeletons and the $m$ new skeletons to minimize the total cost of the assignment. Typically, the cost matrix used for tracking 3D point clouds is constructed by computing the Euclidean distance between the centroids of each point cloud~\cite{weng20203d}. This is more complex with skeletons because the number of points is small and variable. In this context, it is necessary to find a robust cost matrix to make the Hungarian method effective.

First, we build a distance score matrix $D$ based on the Euclidean distance. Given two skeletons $s_{i,k-1}$ and $s_{j,k}$, the average of the distance of common keypoints between $s_{i,k-1}$ and $s_{j,k}$ is stored in $D_{i,j}$.
If two sets of keypoints are disjointed, $D_{i,j}$ is left empty. At the end of the computation, all empty cells are filled with the highest element $max(D)$. At the same time, we build a confidence score matrix $C$. Position $C_{i,j}$ stores the mean absolute distance between the confidences of $s_{i,k-1}$ and $s_{j,k}$. In disjointed sets of keypoints, $C$ follows the same rules applied to $D$.  

Matrices $D$ and $C$ are then combined to obtain matrix $M$ such that:
 \begin{equation}
 M_{i,j} = D_{i,j}+C_{i,j}+u(D_{i,j}+C_{i,j}-\delta)
 \end{equation}
where $u(\cdot)$ is the unit step function and $\delta$ is a fixed step constant. The unit step function increases the assignment cost between skeletons whose combined score is more than $\delta$. 

The cost matrix $M$ is then fed into a linear sum assignment solver to obtain the label assignments that minimize the total cost. Using $M$ instead of $D$ makes the assignment more robust to occlusions and difficult scenarios, such as intersecting people and partially visible body parts. 

Once the assignment is performed, each skeleton $s_j$ is enriched with the cost of its assignment $M_{i,j}$, useful for the permanence filter.

\subsection{Confidence-Based Permanence Filter}
When using a single camera, occlusions often occur during collaborative tasks, undermining the outcome of the interaction. To address this problem, the final node of the proposed pipeline smooths keypoints even during prolonged and complete occlusions. 

The filter core is based on the Object Permanence Filter (OPF)~\cite{Peng2024ObjectPF}, a state-observer originally designed to smooth objects' trajectories.
Similar to a particle filter, the prediction step is defined as:
\begin{align}
    &s^{(i)}_{k+1|k} = f(s^{(i)}_{k|k}, u_k) + \epsilon_k, ~\epsilon_k \sim N(0,Q)
\end{align}
where the state $s = [x, y, z] \in \mathbb{R}^3$ is the cartesian position of the keypoint, $s^{(i)}_{k+1|k}$ is the state of the $i^{th}$ keypoint in timestamp $k+1$ given the past $k$ measurements, $f(\cdot)$ is the state dynamics function, and $Q$ is the covariance of the process noise. 

After the prediction step, if the measurement confidence of the keypoint is higher than the occlusion threshold $T$, then the correction step is performed by the \textit{particle filter update}:
\begin{align}
\label{eq:pf_observation}
w^{(i)}_{k+1|k+1} = \eta P(y_{k+1}|x^{(i)}_{k+1|k})w^{(i)}_{k+1|k}
\end{align}
where $\eta$ is a normalization factor, and $P(y_{k+1}|x^{(i)}_{k+1|k})$ is a Gaussian that represents how the particles approximate the true object state. It is defined as:
\begin{equation}
\label{eq:pf_update}
\resizebox{.85\hsize}{!}{$\begin{aligned}
    P(y_{k+1}|x^{(i)}_{k+1|k})= \frac{1}{\sqrt{(2\pi)^p \det(R)}} \exp\left(-\frac{\nu^T_{k+1}R^{-1}\nu_{k+1}}{2}\right)
\end{aligned}$}
\end{equation}
where 
$\nu_k$ is a Gaussian observation noise and $R$ is the measurement noise covariance matrix. Due to the probabilistic nature of neural networks, the noise level of the keypoints may significantly change. The proposed filter is aware of this, so it changes $R$ dynamically, depending on the keypoint confidence:
\begin{equation}
R_{i,k} = \alpha^{c_{i,k}^{}-\beta}
\end{equation}
where $\alpha$ and $\beta$ are fixed parameters. In this case, when the measurement confidence is low, the covariance is high, resulting in a larger search area; if the measurement confidence is high, the covariance is low, which indicates a more conservative state update.
If a keypoint does not have any measurement available, or the confidence of the measurement is below the occlusion threshold $T$, the \textit{dynamics module} models the dynamics of the keypoint under occlusion. This module approximates the motion under occlusion by fitting a first-order polynomial using the stored length trajectory $\gamma$.


At the end of this phase, the skeletons are refined and transmitted to the \textit{tracking target}, a node that selects which wrist to follow and sends its position to the velocity controller.


\section{Experiments}

\subsection{Experimental Setup}  
We chose the Intel Realsense D455 for the RGB-D camera, providing $848\times480@30$ Hz matrices on both RGB and depth streams. For the HPE backbone, we used OpenPose~\cite{cao2017realtime}, one of the most known 2D multi-person HPEs. The 3D skeletons are obtained by re-projecting the 2D poses into the 3D camera space, as in~\cite{martini2022enabling}. We tracked the operator's left wrist using a marker-based motion capture system (OptiTrack), considered ground truth. We calibrated both the robot and the camera with the Kabsch algorithm~\cite{kabsch1976solution}, aligning them with the frame defined by the OptiTrack system. The experiment runs on 2 PCs with an Intel 12700K and RTX 3090 GPU. One controls the robot and the camera stream, while the other runs the HPE backbone and the filters. Each block in Figure~\ref{fig:pipeline} is implemented as a ROS node and runs in real-time at $30$ Hz.
For the robot, we used the \textit{Franka Emika Panda}, a 7-DOF robotic arm designed for collaborative tasks. 
We employ a passive velocity controller $\tau_c$ as defined in~\cite{7358081}.  Given the joint state variable $\xi$, with velocity $\Dot{\xi}$, and acceleration $\Ddot{\xi}$, the robot has the following rigid-body dynamics:
\begin{align}
\label{eq:passive_dynamics}
M(\xi)\Ddot{\xi} + C(\xi, \Dot{\xi})\xi + g(\xi) = \tau_c + \tau_e
\end{align}
where $M(\cdot)$ is the state-dependent inertia matrix, $C(\cdot)$ is a matrix that captures the Coriolis forces,
$g(\cdot)$  is the gravity vector, and $\tau_e$ refers to external forces interacting with the robot.
The passive velocity controller is defined as follows:
\begin{align}
\label{eq:passive_ds}
\tau_c &= g(\xi) - D(\xi)(\Dot{\xi} - f(\xi^*)), & \Dot{\xi}^* &= f(\xi^*)
\end{align}
where $D(\cdot)$ is a damping term to induce passivity, and $f(\cdot)$ is a simple linear dynamic system tracking the target $\xi^*$ recieved from the filtering pipeline. 
This controller formulation allows humans to push the robot without causing an antagonistic reaction, \update{although no collision occurred}. The HPE backbone, the tracking target node, and the velocity controller are kept fixed during all the experiments.



\subsection{Participants}
A total of 7 adults participated in the experiments (28.5\% female (n=2), 71.5\% male(n=5)). The mean age of the participants was 25.8 years (SD=1.94). The participants were predominantly right-handed, 85.7\% (n=6) and 14.3\% left-handed (n=1). To simulate experienced operators in an industrial plant, all participants were graduate students with a mechanical engineering background who had previously worked with a robotic manipulator.

\subsection{Tasks}
\begin{figure}[t]
    \centering
\includegraphics[width=.826\linewidth]{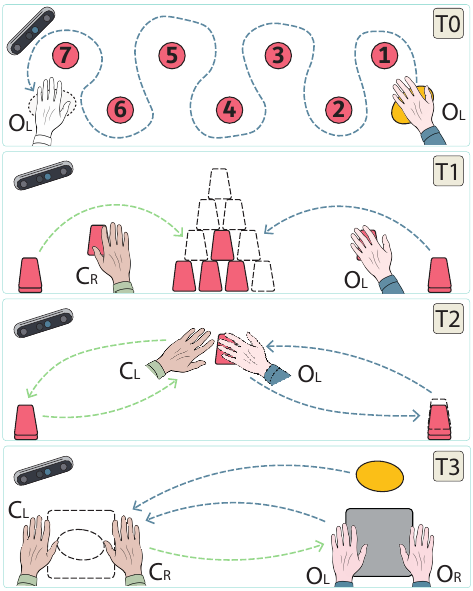}
    \caption{Visual representations of the four tasks.}
    \label{fig:tasks}
\end{figure}

Figure~\ref{fig:tasks} is a graphical representation of the four tasks designed to evaluate the robustness of the proposed pipeline. Each task has an operator $O$, a human collaborator $C$, and a robot $R$. The task of $R$ is to track $O_L$ (i.e., the left hand of $O$) at a pre-defined safety distance of $150$mm  along the $x$ and $z$ axes. 
In particular, the tasks are:
\begin{itemize}
    \item \textbf{T0:} $O_L$ moves a small object from $O$'s right side to the left, avoiding obstacles in a sinusoidal pattern. $C$ is not involved but is still visible during the entire task. This task is designed to test the filter capability in non-linear movements.
    \item \textbf{T1:} $O_L$ and $C_R$ (i.e., $C$'s right hand) collaboratively place 10 cups on each other to form a pyramid. They start with 5 cups each and are not instructed on how to form the pyramid. We designed this task to test the influence of the interaction between $O_L$ and $C_R$.
    \item \textbf{T2:} $O_L$ picks up 5 cups and hands them over to $C_L$, one at a time. Once the first handover is complete, the task is repeated in the other direction (from $C_L$ to $O_L$). This task is designed to test the robustness of the filters when $O_L$ and $C_L$ are very close.
    \item \textbf{T3:} $O$ picks up a box with both $O_L$ and $O_R$, handing it over to $C$. While $C$ opens the box, $O_L$ picks and hands over an object to $C_R$. Then, $C_R$ places the object in the box, $C$ closes it, and hands it back to $O$. We designed this task to test the filter's efficacy in detecting the occlusion of $O_L$ during the heavy handover.
\end{itemize}
\begin{table*}[t!]
\centering
\caption{Tracking quality assessment measuring Mean Absolute Error (MAE), Standard Deviation of the Euclidean Distance (STD), and Mean Acceleration Error (ACC). \update{The average computation time of each filter is expressed in \textit{ms}.}}
\label{tab:sensing}
\resizebox{\linewidth}{!}{%
\begin{tabular}{l|ccc|ccc|ccc|ccc|c}
\toprule
Tasks & \multicolumn{3}{c|}{\textbf{T0}} & \multicolumn{3}{c|}{\textbf{T1}} & \multicolumn{3}{c|}{\textbf{T2}} & \multicolumn{3}{c}{\textbf{T3}} & \multicolumn{1}{|c}{\update{avg}} \\
Evaluation Metrics & \begin{tabular}[c]{@{}c@{}}MAE \\ ($mm$)\end{tabular} & \begin{tabular}[c]{@{}c@{}}STD\\ ($mm$)\end{tabular} & \begin{tabular}[c]{@{}c@{}}ACC\\ ($m/{s^2}$)\end{tabular} & \begin{tabular}[c]{@{}c@{}}MAE \\ ($mm$)\end{tabular} & \begin{tabular}[c]{@{}c@{}}STD\\ ($mm$)\end{tabular} & \begin{tabular}[c]{@{}c@{}}ACC\\ ($m/{s^2}$)\end{tabular} & \begin{tabular}[c]{@{}c@{}}MAE \\ ($mm$)\end{tabular} & \begin{tabular}[c]{@{}c@{}}STD\\ ($mm$)\end{tabular} & \begin{tabular}[c]{@{}c@{}}ACC\\ ($m/{s^2}$)\end{tabular} & \begin{tabular}[c]{@{}c@{}}MAE \\ ($mm$)\end{tabular} & \begin{tabular}[c]{@{}c@{}}STD\\ ($mm$)\end{tabular} & \begin{tabular}[c]{@{}c@{}}ACC\\ ($m/{s^2}$)\end{tabular} & \begin{tabular}[c]{@{}c@{}}\update{time}\\ ($ms$)\end{tabular} \\
\midrule
Baseline & 66.69 & 163.15 & 12.65 & 195.07 & 271.69 & 11.83 & 147.39 & 173.12 & 11.88 & 319.25 & 257.59 & 11.72 & - \\
Kalman Filter (1st) & 69.57 & 127.88 & 6.94 & 149.52 & 215.98 & \textbf{6.43} & 227.56 & 192.59 & 7.27 & 315.66 & 262.06 & 8.20 & 2.95 \\
Kalman Filter (2nd) & 64.19 & 122.18 & \textbf{6.63} & 163.28 & 192.8 & 6.75 & 149.35 & 147.12 & 6.54 & 258.13 & 241.96 & 7.71 & 3.01 \\
\textbf{Ours} & \textbf{48.89} & \textbf{19.55} & 7.24 & \textbf{96.98} & \textbf{79.37} & 8.86 & \textbf{81.62} & \textbf{68.12} & \textbf{6.07} & \textbf{77.47} & \textbf{83.83} & \textbf{7.57} & 8.22\\
\bottomrule
\end{tabular}%
}
\end{table*}
\begin{figure*}[t]
\centering
\begin{subfigure}{0.5\columnwidth}
    \centering
    \includegraphics[width=\columnwidth]{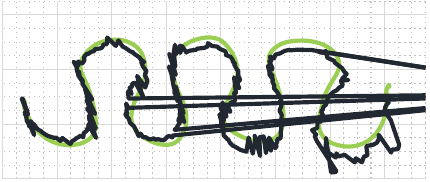}
    \caption{Baseline}
\end{subfigure}%
\begin{subfigure}{0.5\columnwidth}
    \centering
    \includegraphics[width=\columnwidth]{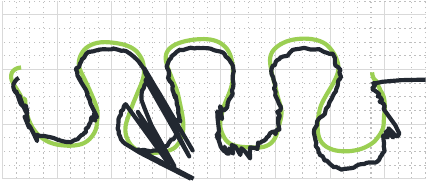}
    \caption{Kalman Filter (1st)}
\end{subfigure}%
\begin{subfigure}{0.5\columnwidth}
    \centering
    \includegraphics[width=\columnwidth]{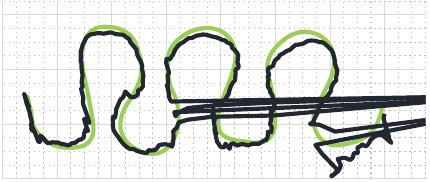}
    \caption{Kalman Filter (2nd)}
\end{subfigure}%
\begin{subfigure}{0.5\columnwidth}
    \centering
    \includegraphics[width=\columnwidth]{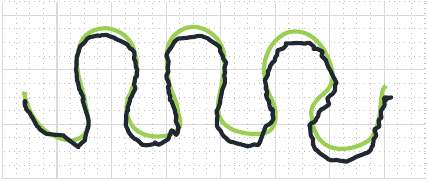}
    \caption{\textbf{Ours}}
\end{subfigure}%
\caption{2D projection on the XY axis of the task T0. In green is the wrist position captured by the ground truth, in black the compared marker-less methods: OpenPose \cite{cao2017realtime} without filters (a), OpenPose filtered by Kalman filter of the first order (b), second order (c) and our filter (d). } 

\label{fig:qualitative}
\end{figure*}
\subsection{Baselines and implementation details}
We compared our filter with two widely used implementations of the Linear Kalman filter. 
The first motion model follows first-order dynamics (\textit{1st}) with constant velocity. The second motion model follows second-order dynamics (\textit{2nd}) with constant acceleration. The rest of the parameters are set as in~\cite{MathWorks_2023}. 

We found the optimal values for each parameter \update{through grid search} and then fixed them throughout the experiments.
In the spatial evaluation node, the distance threshold $\varDelta$ is set to $3.0$ m, and both the reward factor $r$ and the penalty function $p$ are set to $10\%$. The step constant $\delta$ is set to $0.5$ in the temporal evaluation node.
In the permanence filter, we assume first-order polynomial dynamics for $f$. For the observation covariance noise $R$, $\alpha$ is set to $0.01$ and $\beta$ is set to $0.2$. The occlusion threshold $T$ is set to $0.4$, and the size of the dynamic module's past trajectory $\gamma$ is set to $50$ frames.

\subsection{Evaluation metrics}
We compared the output of the \textit{tracking target node} with the ground truth obtained from the marker-based motion capture system. We quantified the quality of the tracking with three metrics:
\begin{itemize}
    \item \textbf{MAE}: it is the average of the Euclidean distances between the two 3D trajectories
    \item \textbf{STD}: it is the standard deviation of the Euclidean distances between the two 3D trajectories
    \item \textbf{ACC}: it is the average of the Euclidean distances between the second derivatives of the two 3D trajectories
\end{itemize} 




We also collected the end-effector's position to quantify compliance with the predefined safety distance, defined as the standard deviation of the Euclidean distances between $O_L$ and the end-effector. We collected participants' feedback through a survey.
Before beginning the study, we did not inform the operator about the filtering methods. For each task, we ran the compared methods in random order. After each task, participants rated the following question: \textit{``How safe did you feel during this task?"} on a Likert scale to quantify the perceived safety. \update{
We performed a one-way ANOVA 
 and then the Tukey-Kramer test to identify statistically significant differences between the methods' mean ratings.
}


\section{Results}
Table~\ref{tab:sensing} compares our filter with the baseline and the other filters.
On average, our method improved the MAE of the baseline by $49.33\%$ ($105.86$mm), outperforming 1st and 2nd order Kalman filters by $51.11\%$ ($114.33$mm) and $44.94\%$ ($82.49$mm), respectively. Also, STD was reduced similarly. On average, our approach showed a $71.72\%$ ($153.67$mm) decrease in STD w.r.t. the baseline, and outperformed the 1st and 2nd order Kalman filters by $70.15\%$ ($136.91$mm) and $65.47\%$ ($13.29$mm), respectively. We then observed a $38.04\%$ ($4.58$ m/s$^2$) reduction in ACC compared to the baseline, achieving comparable performance to Kalman filters designed to provide smooth behavior. \update{On average, our method's computation time is negligible and comparable to the other tested methods.}
\begin{table}[t]
\centering
\caption{Safety distance compliance measuring the standard deviation of the distance between end-effector and wrist.}
\label{tab:ee_std}
\begin{tabular}{lcccc}
\toprule
\textbf{Tasks} & \textbf{T0} & \textbf{T1} & \textbf{T2} & \textbf{T3} \\
\midrule
Baseline & 171.42 & 161.72 & 95.37 & 121.23 \\
Kalman Filter (1st) & 139.4 & 115.71 & 98.65 & 119.24 \\
Kalman Filter (2nd) & 144.5 & 104.17 & 80.08 & 103.91 \\
\textbf{Ours} & \textbf{118.84} & \textbf{46.58} & \textbf{44.66} & \textbf{86.78} \\
\midrule
Marker-based & 113.64 & 33.99 & 42.4 & 77.93\\
\bottomrule
\end{tabular}%
\end{table}

Table~\ref{tab:ee_std} shows the standard deviation of the distance between the robot end-effector and $O_L$. From this perspective, our filter is almost as consistent as the ground truth (only $14.57 \%$ ($7.22$mm)  higher). Table~\ref{tab:likert} assesses participants' perceptions of safety during the experiments. The results indicated a notable increase in perceived safety and comfort when interacting with the robot using our proposed method ($4.6/5$ vs $2.2/5$). \update{
The Tukey-Kramer test proved statistically significant differences between our method and the others.
}
\begin{table}[h]
\centering
\caption{Perceived safety by the participants on a Likert scale \update{and statistical difference from the baseline ($\text{pval} < 0.05$), along with Lower Confidence Interval (LCI) and Upper Confidence Interval (UCI)}.}
\label{tab:likert}
\begin{tabular}{lccccc}
\toprule
\textbf{Tasks} & \textbf{T0} & \textbf{T1} & \textbf{T2} & \textbf{T3} & \textbf{\update{diff (LCI, UCI)}}\\
\midrule
Baseline & 2.1 & 1.2 & 1.8 & 1.7 & -\\
Kalman Filter (1st) & 2.6 & 2.3 & 2.7 & 2.2 & +1.3 (+0.1, +0.7)\\
Kalman Filter (2nd) & 3.1 & 2.3 & 2.7 & 2.3 & +1.5 (+0.2, +0.8)\\
\textbf{Ours} & \textbf{4.7} & \textbf{4.3} & \textbf{4.7} & \textbf{4.5} & \textbf{+3.5 (+2.1, +2.8)} \\
\bottomrule
\end{tabular}%
\end{table}


For qualitative inspection, Figure~\ref{fig:qualitative} shows the 2D projection of $O_L$ during task T0. By minimizing the jitter and large errors caused by occlusions and false predictions and tracking errors, our method produces qualitatively superior results. Since the trajectories of the other three tasks are too complex to represent in 2D, we provide a qualitative video in the supplementary materials.



\section{Conclusion}
Traditional marker-based motion capture systems, while offering high accuracy, are often impractical for everyday use due to their overhead and cost. Markerless motion tracking offers a promising solution for more natural and unobtrusive interactions. However, existing methods face accuracy challenges when dealing with complex scenarios such as occlusions and multi-person interactions. To address these limitations, we proposed a novel filtering pipeline that improved the reliability of any HPE in multi-person HRI environments. Our experimental results demonstrated the effectiveness of the filter in various real-world tasks, outperforming traditional tracking algorithms in terms of safety, smoothness, and user comfort.

\bibliographystyle{IEEEtran}
\bibliography{bibliography}

\end{document}